\documentclass[10pt,twocolumn,letterpaper]{article}

\usepackage{cvpr}
\usepackage{times}
\usepackage{epsfig}
\usepackage{graphicx}
\usepackage{amsmath}
\usepackage{amssymb}

\usepackage[breaklinks=true,bookmarks=false]{hyperref}

\cvprfinalcopy 

\def\cvprPaperID{1726} 

\ifcvprfinal\fi
\begin{document}

\title{MS-TCN: Multi-Stage Temporal Convolutional Network for Action Segmentation}

\author{Yazan Abu Farha and Juergen Gall\\
University of Bonn, Germany\\
{\tt\small \{abufarha,gall\}@iai.uni-bonn.de}
}

\maketitle

\begin{abstract}
   Temporally locating and classifying action segments in long untrimmed 
videos is of particular interest to many applications like surveillance 
and robotics. While traditional approaches follow a two-step pipeline, 
by generating frame-wise probabilities and then feeding them to high-level 
temporal models, recent approaches use temporal convolutions to directly 
classify the video frames. In this paper, we introduce a multi-stage 
architecture for the temporal action segmentation task. Each stage features 
a set of dilated temporal convolutions to generate an initial prediction that 
is refined by the next one. This architecture is trained using a combination 
of a classification loss and a proposed smoothing loss that penalizes 
over-segmentation errors. Extensive evaluation shows the effectiveness of the 
proposed model in capturing long-range dependencies and recognizing action 
segments. Our model achieves state-of-the-art results on three challenging 
datasets: 50Salads, Georgia Tech Egocentric Activities (GTEA), and the Breakfast dataset.
\end{abstract}

\section{Introduction}
Analyzing activities in videos is of significant importance for 
many applications ranging from video indexing to surveillance. 
While methods for classifying short trimmed videos have been very 
successful~\cite{carreira2017quo, feichtenhofer2016spatiotemporal}, 
detecting and temporally locating action segments in long untrimmed 
videos is still challenging.

Earlier approaches for action segmentation can be grouped into two categories: 
sliding window approaches~\cite{rohrbach2012database, karaman2014fast, 
oneata2014lear}, that use temporal windows of different scales to detect 
action segments, and hybrid approaches that apply a coarse temporal modeling using 
Markov models on top of frame-wise classifiers~\cite{kuehne2016end, lea2016segmental, 
richard2017weakly}. While these approaches achieve good results, they are very 
slow as they require solving a maximization problem over very long sequences.

Motivated by the advances in speech synthesis, recent approaches rely on 
temporal convolutions to capture long range dependencies between the video 
frames~\cite{Lea_2017_CVPR, lei2018temporal, ding2018weakly}. In these models, 
a series of temporal convolutions and pooling layers are adapted in an 
encoder-decoder architecture for the temporal action segmentation. Despite 
the success of such temporal models, these approaches operate on a very low 
temporal resolution of a few frames per second.

\begin{figure}[tb]
\begin{center}
   \includegraphics[width=.95\linewidth]{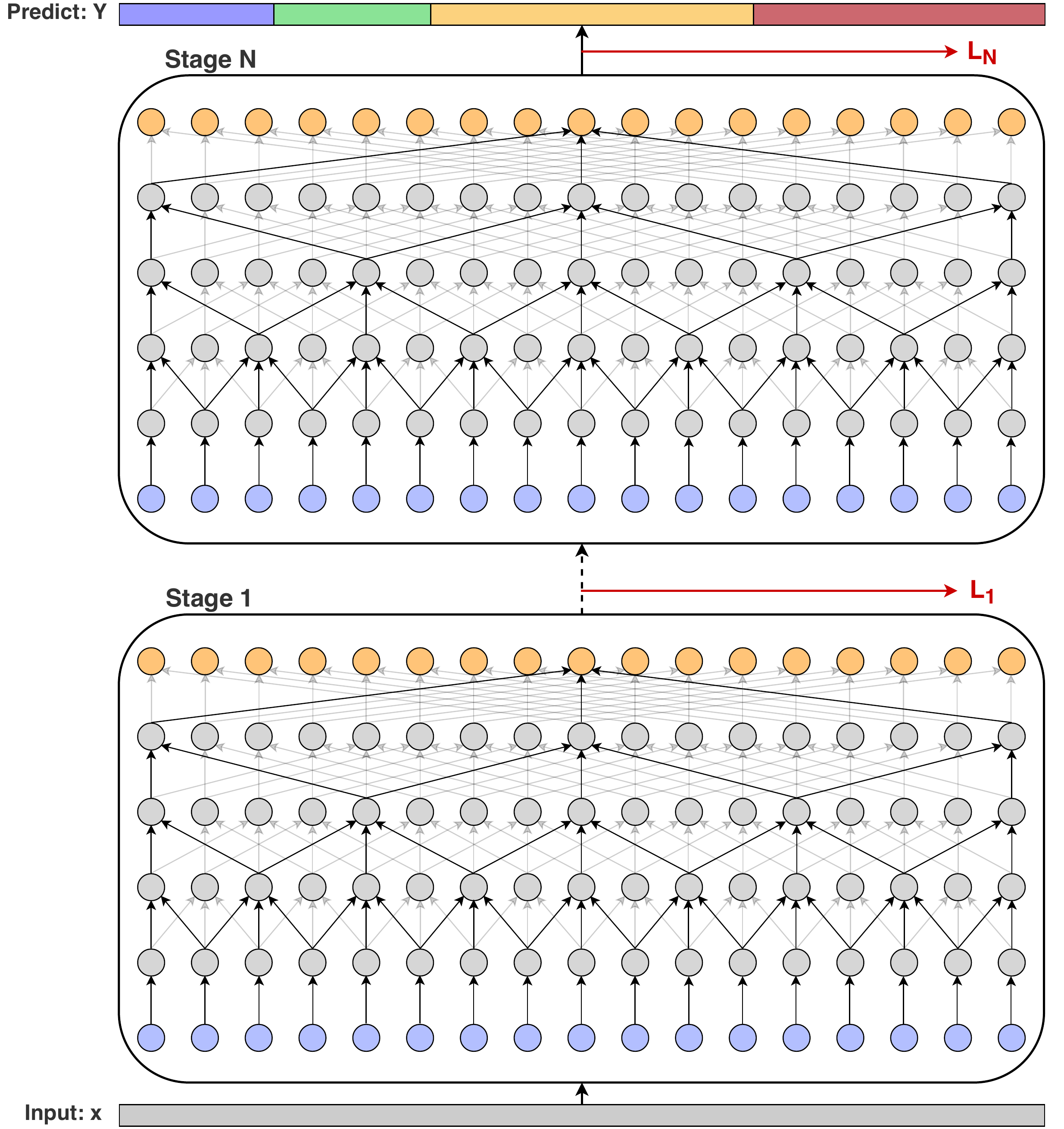}
\end{center}
   \caption{Overview of the multi-stage temporal convolutional network. 
   Each stage generates an initial prediction that is refined by the next stage. 
   At each stage, several dilated 1D convolutions are applied on the activations 
   of the previous layer. A loss layer is added after each stage.}
\label{fig:model}
\end{figure}

In this paper, we propose a new model that also uses temporal convolutions 
which we call Multi-Stage Temporal Convolutional Network (MS-TCN). 
In contrast to previous approaches, the proposed model operates 
on the full temporal resolution of the videos and thus achieves better 
results. Our model consists of multiple stages where each stage outputs 
an initial prediction that is refined by the next one. In each stage, we 
apply a series of dilated 1D convolutions, which enables the model to 
have a large temporal receptive field with less parameters. Figure~\ref{fig:model} 
shows an overview of the proposed multi-stage model. While this architecture 
already performs well, we further employ a smoothing loss during training 
which penalizes over-segmentation errors in the predictions. Extensive 
evaluation on three datasets shows the effectiveness of our model in capturing 
long range dependencies between action classes and producing high quality 
predictions. Our contribution is thus two folded: First, we propose a multi-stage 
temporal convolutional architecture for the action segmentation task that operates on 
the full temporal resolution. Second, we introduce a smoothing loss to enhance the 
quality of the predictions. Our approach achieves state-of-the-art results on three 
challenging benchmarks for action segmentation: 50Salads~\cite{stein2013combining}, 
Georgia Tech Egocentric Activities (GTEA)~\cite{fathi2011learning}, and the Breakfast 
dataset~\cite{kuehne2014language}. \footnote{The source code for our model is publicly 
available at \url{https://github.com/yabufarha/ms-tcn}.}

\section{Related Work}

Detecting actions and temporally segmenting long untrimmed videos has been 
studied by many researchers. While traditional approaches use a sliding window 
approach with non-maximum suppression~\cite{rohrbach2012database, karaman2014fast}, 
Fathi and Rehg~\cite{fathi2013modeling} model actions based on the change in the state 
of objects and materials. In~\cite{fathi2011understanding}, actions are represented 
based on the interactions between hands and objects. These representations are used to 
learn sets of temporally-consistent actions. Bhattacharya \etal~\cite{bhattacharya2014recognition} 
use a vector time series representation of videos to model the temporal dynamics of complex 
actions using methods from linear dynamical systems theory. The representation is based on 
the output of pre-trained concept detectors applied on overlapping temporal windows. 
Cheng \etal~\cite{cheng2014temporal} represent videos as a sequence of visual 
words, and model the temporal dependency by employing a Bayesian non-parametric model 
of discrete sequences to jointly classify and segment video sequences.

Other approaches employ high level temporal modeling over frame-wise classifiers.  
Kuehne \etal~\cite{kuehne2016end} represent the frames of a video using Fisher vectors of 
improved dense trajectories, and then each action is modeled with a hidden Markov 
model (HMM). These HMMs are combined with a context-free grammar for recognition 
to determine the most probable sequence of actions. A hidden Markov model is also 
used in~\cite{tang2012learning} to model both transitions between states and their durations.
Vo and Bobick~\cite{vo2014stochastic} use a Bayes network to segment activities. 
They represent compositions of actions using a stochastic context-free grammar with 
AND-OR operations.
\cite{richard2016temporal} propose a model for temporal action detection that 
consists of three components:  an action model that maps features extracted from 
the video frames into action probabilities, a language model that describes the 
probability of actions at sequence level, and finally a length model that models 
the length of different action segments. To get the video segmentation, they use 
dynamic programming to find the solution that maximizes the joint probability of 
the three models.
Singh \etal~\cite{singh2016multi} use a two-stream network to learn representations 
of short video chunks. These representations are then passed to a bi-directional 
LSTM to capture dependencies between different chunks. However, their approach is 
very slow due to the sequential prediction.
In~\cite{singh2016first}, a three-stream architecture that operates on spatial, 
temporal and egocentric streams is introduced to learn  egocentric-specific features. 
These features are then classified using a multi-class SVM.

Inspired by the success of temporal convolution in speech synthesis~\cite{van2016wavenet}, 
researchers have tried to use similar ideas for the temporal action segmentation 
task. Lea \etal~\cite{Lea_2017_CVPR} propose a temporal convolutional network for action 
segmentation and detection. Their approach follows an encoder-decoder architecture with 
a temporal convolution and pooling in the encoder, and upsampling followed by deconvolution 
in the decoder. While using temporal pooling enables the model to capture long-range dependencies, 
it might result in a loss of fine-grained information that is necessary for fine-grained recognition. 
Lei and Todorovic~\cite{lei2018temporal} build on top of~\cite{Lea_2017_CVPR} and use deformable 
convolutions instead of the normal convolution and add a residual stream to the encoder-decoder model. 
Both approaches in ~\cite{Lea_2017_CVPR, lei2018temporal} operate on downsampled videos with 
a temporal resolution of 1-3 frames per second.
In contrast to these approaches, we operate on the full temporal resolution and 
use dilated convolutions to capture long-range dependencies.

There is a huge line of research that addresses the action segmentation task in 
a weakly supervised setup \cite{bojanowski2014weakly, huang2016connectionist, kuehne2017weakly, 
richard2017weakly, ding2018weakly}. 
Kuehne \etal~\cite{kuehne2017weakly} train a model for action segmentation from 
video transcripts. In their approach, an HMM is learned for each action and a Gaussian 
mixture model (GMM) is used to model observations. However, since frame-wise classifiers 
do not capture enough context to detect action classes, Richard \etal~\cite{richard2017weakly} 
use a GRU instead of the GMM that is used in~\cite{kuehne2017weakly}, and they further 
divide each action into multiple sub-actions to better detect complex actions. 
Both of these models are trained in an iterative procedure starting from a linear alignment 
based on the video transcript. Similarly, Ding and Xu~\cite{ding2018weakly} train a temporal 
convolutional feature pyramid network in an iterative manner starting from a linear alignment. 
Instead of using hard labels, they introduce a soft labeling mechanism at the 
boundaries, which results in a better convergence. In contrast to these approaches, we address 
the temporal action segmentation task in a fully supervised setup and the weakly 
supervised case is beyond the scope of this paper.

\section{Temporal Action Segmentation}
We introduce a multi-stage temporal convolutional network for the temporal
action segmentation task. Given the frames of a video 
$x_{1:T} = (x_1,\dots,x_T)$, our goal is to infer the class label for 
each frame $c_{1:T} = (c_1,\dots,c_T)$, where $T$ is the video length.  
First, we describe the single-stage approach in 
Section~\ref{sec:single_stage_model}, then we discuss the multi-stage 
model in Section~\ref{sec:multi_stage_model}. Finally, we describe the 
proposed loss function in Section~\ref{sec:loss_function}.
 
\subsection{Single-Stage TCN}
\label{sec:single_stage_model}
\begin{figure}[tb]
\begin{center}
   \includegraphics[width=0.4\linewidth]{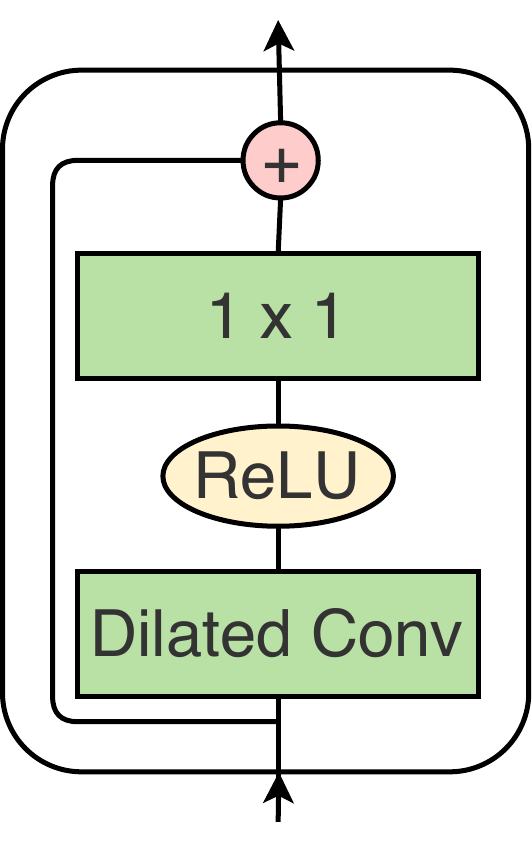}
\end{center}
   \caption{Overview of the dilated residual layer.}
\label{fig:layer}
\end{figure}

Our single stage model consists of only temporal convolutional layers. 
We do not use pooling layers, which reduce the temporal resolution, or 
fully connected layers, which force the model to operate on inputs of 
fixed size and massively increase the number of parameters. We call this 
model a single-stage temporal convolutional network (SS-TCN).
The first layer of a single-stage TCN is a $1 \times 1$ convolutional layer, 
that adjusts the dimension of the input features to match the number 
of feature maps in the network. Then, this layer is followed by several 
layers of dilated 1D convolution. Inspired by the wavenet~\cite{van2016wavenet} 
architecture, we use a dilation factor that is doubled at each layer, 
\ie $1, 2, 4, ...., 512$. All these layers have the same number of 
convolutional filters. However, instead of the causal convolution that 
is used in wavenet, we use acausal convolutions with kernel size 3. 
Each layer applies a dilated convolution with ReLU activation to the output 
of the previous layer. We further use residual connections to facilitate 
gradients flow. The set of operations at each layer can be formally 
described as follows
 \begin{align}
& \hat{H}_l = ReLU(W_1 * H_{l-1} + b_1), \\
& H_l = H_{l-1} + W_2 * \hat{H}_l + b_2, 
\end{align}
where $H_l$ is the output of layer $l$, $*$ denotes the convolution operator, 
$W_1 \in R^{3 \times D \times D}$ are the weights of the dilated convolution 
filters with kernel size 3 and $D$ is the number of convolutional filters, 
$W_2 \in R^{1 \times D \times D}$ are the weights of a $1 \times 1$ convolution, 
and $b_1, b_2 \in R^{D}$ are bias vectors. These operations are illustrated in 
Figure~\ref{fig:layer}. Using dilated convolution increases the receptive field 
without the need to increase the number of parameters by increasing the number 
of layers or the kernel size. Since the receptive field grows exponentially with 
the number of layers, we can achieve a very large receptive field with a few layers, 
which helps in preventing the model from over-fitting the training data. The receptive 
field at each layer is determined using this formula 
\begin{equation}
ReceptiveField(l) = 2^{l+1} - 1, 
\end{equation}
where $l \in \left[ 1, L\right] $ is the layer number. Note that this formula is only 
valid for a kernel of size 3. To get the probabilities for the output class, we apply a 
$1 \times 1$ convolution over the output of the last dilated convolution layer followed 
by a softmax activation, \ie
\begin{equation}
Y_t = Softmax(Wh_{L,t} + b), 
\end{equation}
where $Y_t$ contains the class probabilities at time $t$, $h_{L,t}$ is the output 
of the last dilated convolution layer at time $t$, $W \in R^{C \times D}$ and 
$b \in R^{C}$ are the weights and bias for the $1 \times 1$ convolution layer, 
where $C$ is the number of classes and $D$ is the number of convolutional filters.

\subsection{Multi-Stage TCN}
\label{sec:multi_stage_model}
Stacking several predictors sequentially has shown significant 
improvements in many tasks like human pose estimation~\cite{wei2016convolutional, 
newell2016stacked}. The idea of these stacked or multi-stage architectures 
is composing several models sequentially such that each model operates directly on 
the output of the previous one. The effect of such composition is 
an incremental refinement of the predictions from the previous stages. 

Motivated by the success of such architectures, we introduce a multi-stage 
temporal convolutional network for the temporal action segmentation
task. In this multi-stage model, each stage takes an initial prediction from 
the previous stage and refines it. The input of the first stage is the 
frame-wise features of the video as follows
\begin{align}
& Y^0 = x_{1:T}, \\
& Y^s = \mathcal{F}(Y^{s-1}), 
\end{align}
where $Y^s$ is the output at stage $s$ and $\mathcal{F}$ is the single-stage TCN 
discussed in Section~\ref{sec:single_stage_model}. Using such a multi-stage 
architecture helps in providing more context to predict the class label at each 
frame. Furthermore, since the output of each stage is an initial prediction, 
the network is able to capture dependencies between action classes and learn 
plausible action sequences, which helps in reducing the over-segmentation errors.

Note that the input to the next stage is just the frame-wise probabilities without 
any additional features. We will show in the experiments how adding features to the 
input of next stages affects the quality of the predictions.

\subsection{Loss Function}
\label{sec:loss_function}

As a loss function, we use a combination of a classification loss and a smoothing 
loss. For the classification loss, we use a cross entropy loss
\begin{equation}
\mathcal{L}_{cls} = \frac{1}{T}\sum_{t} -log(y_{t,c}), 
\end{equation}
where $y_{t,c}$ is the the predicted probability for the ground truth 
label at time $t$.

While the cross entropy loss already performs well, we found that the predictions 
for some of the videos contain a few over-segmentation errors. To further improve 
the quality of the predictions, we use an additional smoothing loss to reduce such 
over-segmentation errors. For this loss, we use a truncated mean squared error over 
the frame-wise log-probabilities

\begin{equation}\label{eqn:t_mse}
\mathcal{L}_{T-MSE} = \frac{1}{TC}\sum_{t,c}\tilde{\Delta}_{t,c}^2, 
\end{equation}
\begin{equation}
\tilde{\Delta}_{t,c} = 
\begin{cases}
\Delta_{t,c}        &: \Delta_{t,c} \leq \tau\\
\tau                &: otherwise\\
\end{cases}, 
\end{equation}
\begin{equation}
\Delta_{t,c} = \left| log\ y_{t,c} - log\ y_{t-1,c} \right|,
\end{equation}
where $T$ is the video length, $C$ is the number of classes, and $y_{t,c}$ 
is the probability of class $c$ at time $t$. 

Note that the gradients are only computed 
with respect to $y_{t,c}$, whereas $y_{t-1,c}$ is not considered as a function 
of the model's parameters. This loss is similar to the Kullback-Leibler (KL) 
divergence loss where 
\begin{equation}
\mathcal{L}_{KL} = \frac{1}{T}\sum_{t,c} y_{t-1,c} (log\ y_{t-1,c} - log\ y_{t,c}).
\end{equation}
However, we found that the truncated mean squared error ($\mathcal{L}_{T-MSE}$)~\eqref{eqn:t_mse} 
reduces the over-segmentation errors more. We will compare the KL loss and the proposed 
loss in the experiments.

The final loss function for a single stage is a combination of the 
above mentioned losses
\begin{equation}
\mathcal{L}_s = \mathcal{L}_{cls} + \lambda \mathcal{L}_{T-MSE}, 
\end{equation}
where $\lambda$ is a model hyper-parameter to determine the contribution of 
the different losses. Finally to train the complete model, we minimize the 
sum of the losses over all stages
\begin{equation}
\mathcal{L} = \sum_s \mathcal{L}_{s} . 
\end{equation}

\subsection{Implementation Details}
\label{sec:implementation details}
We use a multi-stage architecture with four stages, each stage contains 
ten dilated convolution layers, where the dilation factor is doubled at 
each layer and dropout is used after each layer. We set the number of 
filters to $64$ in all the layers of the model and the filter size is $3$. 
For the loss function, we set $\tau = 4$ and $\lambda = 0.15$. In all experiments, 
we use Adam optimizer with a learning rate of $0.0005$.
\section{Experiments}
\paragraph{Datasets.} We evaluate the proposed model on three challenging 
datasets: 50Salads~\cite{stein2013combining}, Georgia Tech Egocentric Activities 
(GTEA)~\cite{fathi2011learning}, and the Breakfast dataset~\cite{kuehne2014language}.

The \textbf{50Salads} dataset contains 50 videos with $17$ action classes. 
On average, each video contains 20 action instances and is $6.4$ minutes long. 
As the name of the dataset indicates, the videos depict salad preparation 
activities. These activities were performed by $25$ actors where each actor 
prepared two different salads. For evaluation, we use five-fold cross-validation 
and report the average as in~\cite{stein2013combining}. 

The \textbf{GTEA} dataset contains $28$ videos corresponding to 7 different 
activities, like preparing coffee or cheese sandwich, performed by 4 subjects. 
All the videos were recorded by a camera that is mounted on the actor's head. 
The frames of the videos are annotated with $11$ action classes including 
background. On average, each video has 20 action instances. We use cross-validation 
for evaluation by leaving one subject out. 

The \textbf{Breakfast} dataset is the largest among the three datasets with 
$1,712$ videos. The videos were recorded in 18 different kitchens showing 
breakfast preparation related activities. Overall, there are $48$ different 
actions where each video contains $6$ action instances on average. For evaluation, 
we use the standard 4 splits as proposed in~\cite{kuehne2014language} and report 
the average.

For all datasets, we extract I3D~\cite{carreira2017quo} features for the video  
frames and use these features as input to our model. For GTEA and Breakfast datasets
we use the videos temporal resolution at $15$ fps, while for 50Salads we downsampled 
the features from $30$ fps to $15$ fps to be consistent with the other datasets.

\paragraph{Evaluation Metrics.} For evaluation, we report the frame-wise 
accuracy (Acc), segmental edit distance and the segmental F1 score at overlapping 
thresholds $10\%,\ 25\%$ and $50\%$, denoted by $F1$@$\{10,25,50\}$. The overlapping 
threshold is determined based on the intersection over union (IoU) ratio. While the 
frame-wise accuracy is the most commonly used metric for action segmentation, long 
action classes have a higher impact than short action classes on this metric and 
over-segmentation errors have a very low impact. For that reason, we use the segmental 
F1 score as a measure of the quality of the prediction as proposed by~\cite{Lea_2017_CVPR}.

\subsection{Effect of the Number of Stages}

\begin{figure}[tb]
\begin{center}
   \includegraphics[trim={1cm 1.5cm 1cm 2.5cm},clip,width=.85\linewidth]{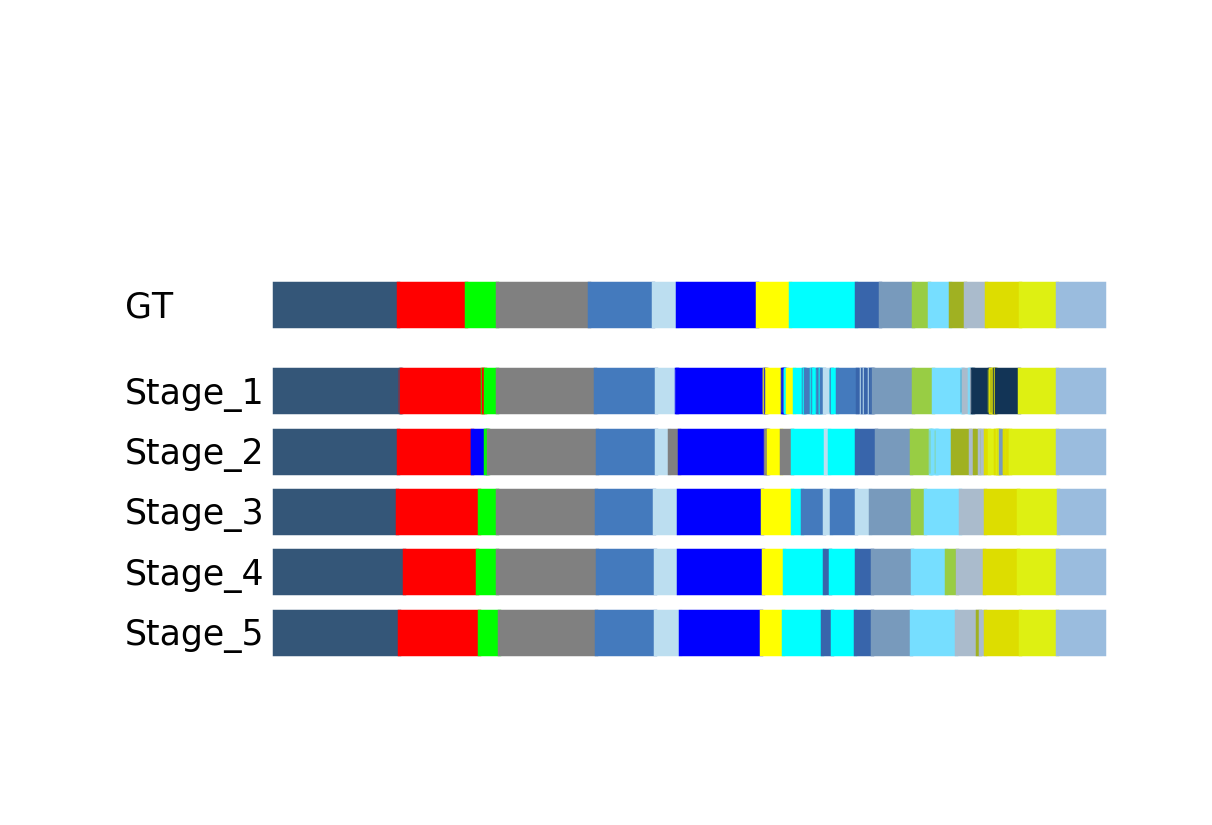}
\end{center}
   \caption{Qualitative result from the 50Salads dataset for comparing different number of stages.}
\label{fig:qualitative_results_stages}
\end{figure}

We start our evaluation by showing the effect of using a multi-stage 
architecture. Table~\ref{tab:number_of_stages} shows the results of a 
single-stage model compared to multi-stage models with different 
number of stages. As shown in the table, all of these models achieve 
a comparable frame-wise accuracy. Nevertheless, the quality of the 
predictions is very different. Looking at the segmental edit distance 
and F1 scores of these models, we can see that the single-stage model 
produces a lot of over-segmentation errors, as indicated by the 
low F1 score. On the other hand, using a multi-stage architecture 
reduces these errors and increases the F1 score. This effect is 
clearly visible when we use two or three stages, which gives a huge 
boost to the accuracy. Adding the fourth stage still improves the 
results but not as significant as the previous stages. However, 
by adding the fifth stage, we can see that the performance starts 
to degrade. This might be an over-fitting problem as a result of 
increasing the number of parameters. The effect of the multi-stage 
architecture can also be seen in the qualitative results shown
in Figure~\ref{fig:qualitative_results_stages}. Adding more stages 
results in an incremental refinement of the predictions. For the rest 
of the experiments we use a multi-stage TCN with four stages.

\begin{table}[tb]
\centering
\resizebox{.75\linewidth}{!}{%
\begin{tabular}{lccccc}
\hline
 & \multicolumn{3}{c}{F1@\{10,25,50\}} & Edit & Acc  
\\ \hline
SS-TCN            &         27.0  &         25.3  &         21.5  &         20.5  &         78.2  \\
MS-TCN (2 stages) &         55.5  &         52.9  &         47.3  &         47.9  &         79.8  \\
MS-TCN (3 stages) &         71.5  &         68.6  &         61.1  &         64.0  &         78.6  \\ 
MS-TCN (4 stages) &         76.3  & \textbf{74.0} & \textbf{64.5} &         67.9  & \textbf{80.7} \\ 
MS-TCN (5 stages) & \textbf{76.4} &         73.4  &         63.6  & \textbf{69.2} &         79.5  \\   
\hline

\end{tabular}%
}
\caption{Effect of the number of stages on the 50Salads dataset.}
\label{tab:number_of_stages}
\end{table}

\subsection{Multi-Stage TCN vs. Deeper Single-Stage TCN}

In the previous section, we have seen that our multi-stage architecture is 
better than a single-stage one. However, that comparison does not show whether 
the improvement is because of the multi-stage architecture or due to the increase 
in the number of parameters when adding more stages. For a fair comparison, we 
train a single-stage model that has the same number of parameters as the multi-stage one. 
As each stage in our MS-TCN contains $12$ layers (ten dilated convolutional layers, one 
$1 \times 1$ convolutional layer and a softmax layer), we train a single-stage TCN with 
$48$ layers, which is the number of layers in a MS-TCN with four stages. For the dilated convolutions, 
we use similar dilation factors as in our MS-TCN. \Ie. we start with a dilation factor 
of $1$ and double it at every layer up to a factor of $512$, and then we start again from 
$1$. As shown in Table~\ref{tab:more_layers}, our multi-stage architecture outperforms 
its single-stage counterpart with a large margin of up to $27\%$. This highlights the 
impact of the proposed architecture in improving the quality of the predictions.

\begin{table}[tb]
\centering
\resizebox{.75\linewidth}{!}{%
\begin{tabular}{lccccc}
\hline
  & \multicolumn{3}{c}{F1@\{10,25,50\}} & Edit & Acc  
\\ \hline
SS-TCN (48 layers) &         49.0  &         46.4  &         40.2  &         40.7  &         78.0  \\
MS-TCN             & \textbf{76.3} & \textbf{74.0} & \textbf{64.5} & \textbf{67.9} & \textbf{80.7} \\ 
\hline
\end{tabular}%
}
\caption{Comparing a multi-stage TCN with a deep single-stage TCN on the 50Salads dataset.}
\label{tab:more_layers}
\end{table}

\subsection{Comparing Different Loss Functions}

\begin{table}[tb]
\centering
\resizebox{.75\linewidth}{!}{%
\begin{tabular}{lccccc}
\hline
  & \multicolumn{3}{c}{F1@\{10,25,50\}} & Edit & Acc  
\\ \hline
$\mathcal{L}_{cls} $       							 &        71.3 &        69.7 &        60.7 &        64.2 &        79.9  \\
$\mathcal{L}_{cls} + \lambda \mathcal{L}_{KL}$	     &        71.9 &        69.3 &        60.1 &        64.6 &        80.2  \\
$\mathcal{L}_{cls} + \lambda \mathcal{L}_{T-MSE}$   &\textbf{76.3}&\textbf{74.0}&\textbf{64.5}&\textbf{67.9}&\textbf{80.7} \\ 
\hline
\end{tabular}%
}
\caption{Comparing different loss functions on the 50Salads dataset.}
\label{tab:loss_function}
\end{table}

\begin{figure}[tb]
\begin{center}
   \includegraphics[trim={.4cm 2.7cm 1cm 2.5cm},clip,width=.85\linewidth]{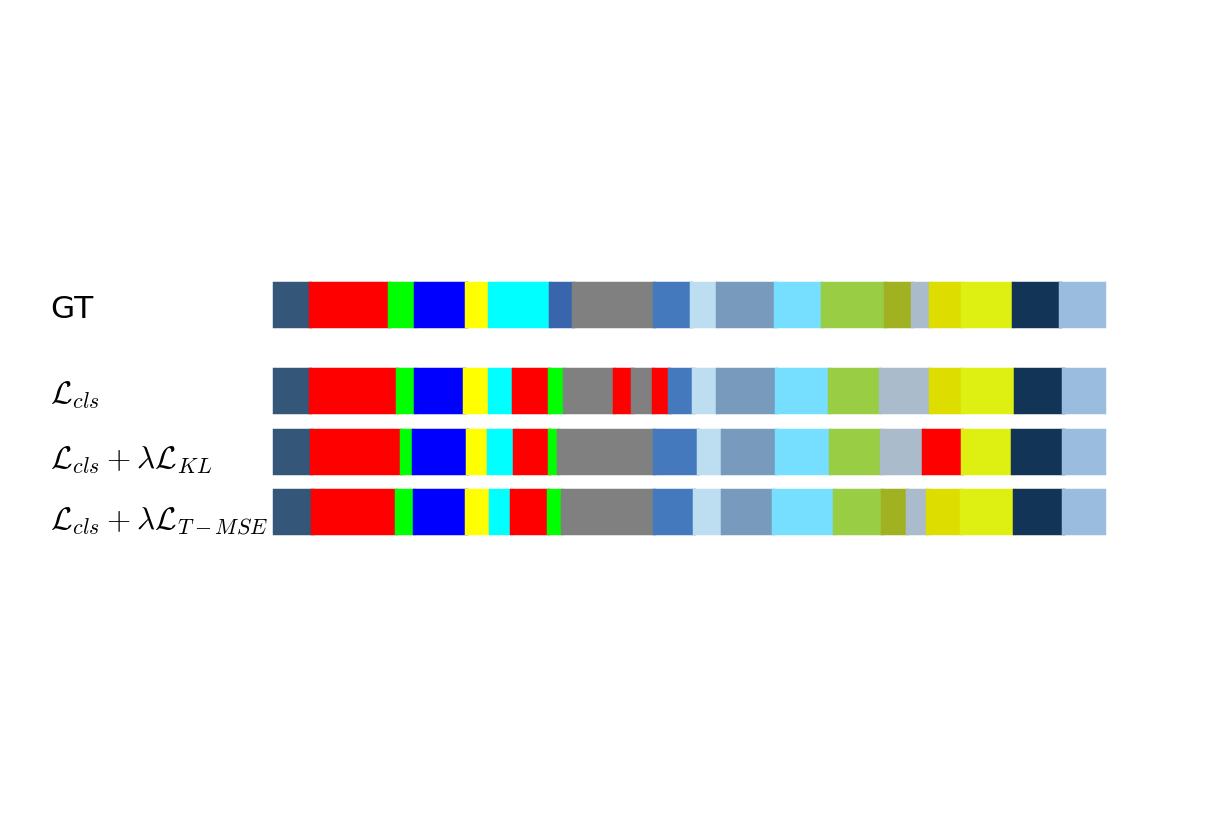}
\end{center}
   \caption{Qualitative result from the 50Salads dataset for comparing different loss functions.}
\label{fig:loss_functions}
\end{figure}

As a loss function, we use a combination of a cross-entropy loss, which 
is common practice for classification tasks, and a truncated mean squared 
loss over the frame-wise log-probabilities to ensure smooth predictions. 
While the smoothing loss slightly improves the frame-wise accuracy compared to 
the cross entropy loss alone, we found that this loss produces much less 
over-segmentation errors. Table~\ref{tab:loss_function} and 
Figure~\ref{fig:loss_functions} show a comparison of these losses. 
As shown in Table~\ref{tab:loss_function}, the proposed loss achieves 
better F1 and edit scores with an absolute improvement of $5\%$. This 
indicates that our loss produces less over-segmentation errors compared 
to cross entropy since it forces consecutive frames to have 
similar class probabilities, which results in a smoother output. 

\begin{figure}[tb]
\begin{center}
\begin{tabular}{cc}
   \includegraphics[trim={4cm .6cm 27cm .3cm},clip,width=.44\linewidth]{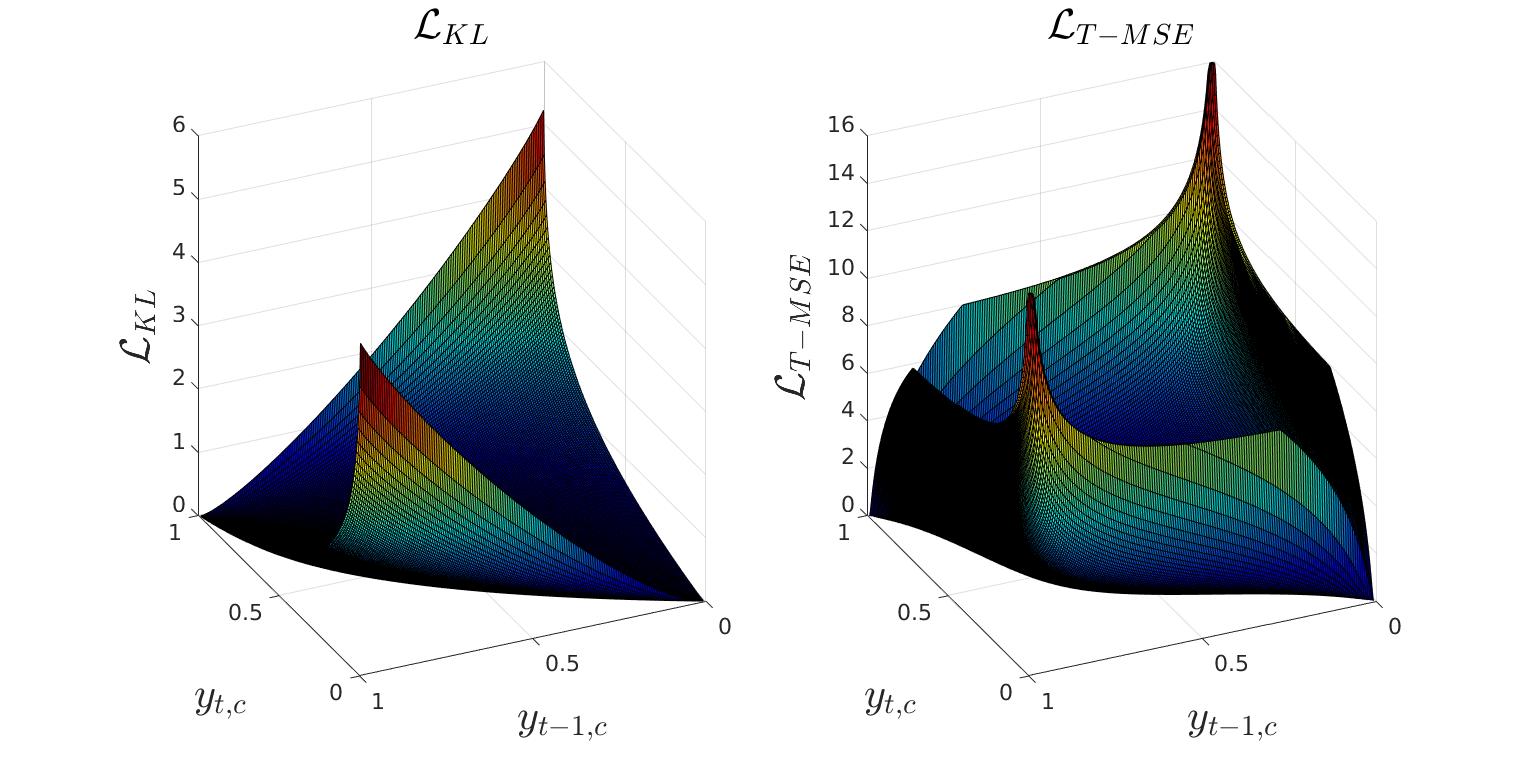} & 
   \includegraphics[trim={27cm .6cm 4cm .3cm},clip,width=.44\linewidth]{losses_plot_new_1}
\end{tabular}
\end{center}
   \caption{Loss surface for the Kullback-Leibler (KL) divergence loss ($\mathcal{L}_{KL}$) 
   and the proposed truncated mean squared loss ($\mathcal{L}_{T-MSE}$) for the case of two 
   classes. $y_{t,c}$ is the predicted probability for class c and $y_{t-1,c}$ is the target 
   probability corresponding to that class.}
\label{fig:losses_plot}
\end{figure}

Penalizing the difference in log-probabilities is similar to the Kullback-Leibler 
(KL) divergence loss, which measures the difference between two probability 
distributions. However, the results show that the proposed loss produces better 
results than the KL loss as shown in Table~\ref{tab:loss_function} 
and Figure~\ref{fig:loss_functions}. The reason behind this is the fact that 
the KL divergence loss does not penalize cases where the difference between the target 
probability and the predicted probability is very small. Whereas the proposed 
loss penalizes small differences as well. Note that, in contrast to the KL loss, 
the proposed loss is symmetric. Figure~\ref{fig:losses_plot} shows the surface 
for both the KL loss and the proposed truncated mean squared loss for the case 
of two classes. We also tried a symmetric version of the KL loss but it performed 
worse than the proposed loss.

\subsection{Impact of $\lambda$ and $\tau$}

The effect of the proposed smoothing loss is controlled by two hyper-parameters: 
$\lambda$ and $\tau$. In this section, we study the impact of these parameters and 
see how they affect the performance of the proposed model.

\noindent\textbf{Impact of $\lambda$:} In all experiments, we set $\lambda=0.15$. To 
analyze the effect of this parameter, we train different models with different 
values of $\lambda$. As shown in Table~\ref{tab:impact_lambda_tau}, the impact 
of $\lambda$ is very small on the performance. Reducing $\lambda$ to $0.05$ 
still improves the performance but not as good as the default value of $\lambda=0.15$. 
Increasing its value to $\lambda=0.25$ also causes a degradation in 
performance. This drop in performance is due to the fact that the smoothing loss 
penalizes heavily changes in frame-wise labels, which affects the detected boundaries 
between action segments.

\noindent\textbf{Impact of $\tau$:} This hyper-parameter defines the threshold to truncate 
the smoothing loss. Our default value is $\tau=4$. While reducing the value to $\tau=3$ 
still gives an improvement over the cross entropy baseline, setting $\tau=5$ results 
in a huge drop in performance. This is mainly because when $\tau$ is too high, the smoothing 
loss penalizes cases where the model is very confident that the consecutive frames 
belong to two different classes, which indeed reduces the capability of the model in 
detecting the true boundaries between action segments.

\begin{table}[tb]
\centering
\resizebox{.75\linewidth}{!}{%
\begin{tabular}{lccccc}
\hline
\textbf{Impact of $\lambda$}  & \multicolumn{3}{c}{F1@\{10,25,50\}} & Edit & Acc  
\\ \hline
MS-TCN ($\lambda=0.05,\ \tau=4$)   &        74.1 &        71.7 &        62.4 &        66.6 &        80.0  \\ 
MS-TCN ($\lambda=0.15,\ \tau=4$)   &\textbf{76.3}&\textbf{74.0}&\textbf{64.5}&        67.9 &\textbf{80.7} \\ 
MS-TCN ($\lambda=0.25,\ \tau=4$)   &        74.7 &        72.4 &        63.7 &\textbf{68.1}&        78.9  \\ 
\hline
\hline
\textbf{Impact of $\tau$}  & \multicolumn{3}{c}{F1@\{10,25,50\}} & Edit & Acc  
\\ \hline
MS-TCN ($\lambda=0.15,\ \tau=3$)   &        74.2 &        72.1 &        62.2 &        67.1 &        79.4  \\ 
MS-TCN ($\lambda=0.15,\ \tau=4$)   &\textbf{76.3}&\textbf{74.0}&\textbf{64.5}&\textbf{67.9}&\textbf{80.7} \\ 
MS-TCN ($\lambda=0.15,\ \tau=5$)   &        66.6 &        63.7 &        54.7 &        60.0 &        74.0  \\ 
\hline
\end{tabular}%
}
\caption{Impact of $\lambda$ and $\tau$ on the 50Salads dataset.}
\label{tab:impact_lambda_tau}
\end{table}

\subsection{Effect of Passing Features to Higher Stages}

\begin{table}[tb]
\centering
\resizebox{.75\linewidth}{!}{%
\begin{tabular}{lccccc}
\hline
  & \multicolumn{3}{c}{F1@\{10,25,50\}} & Edit & Acc  
\\ \hline
Probabilities and features 	&        56.2 &        53.7 &        45.8 &        47.6 &       76.8  \\
Probabilities only          &\textbf{76.3}&\textbf{74.0}&\textbf{64.5}&\textbf{67.9}&\textbf{80.7} \\
\hline
\end{tabular}%
}
\caption{Effect of passing features to higher stages on the 50Salads dataset.}
\label{tab:features}
\end{table}

\begin{figure}[tb]
\begin{center}
   \includegraphics[trim={.4cm 3.5cm 1cm 2.5cm},clip,width=.85\linewidth]{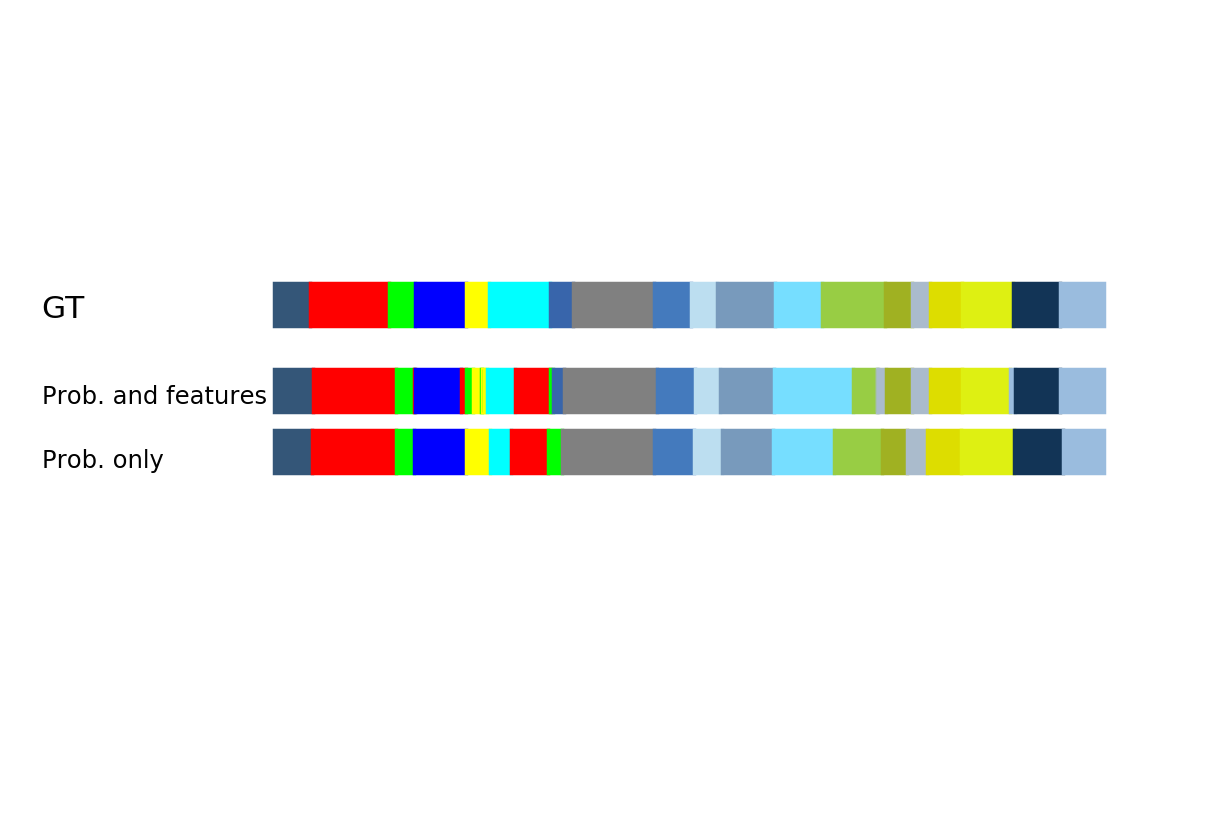}
   \\
   \includegraphics[trim={.4cm 3.5cm 1cm 2.5cm},clip,width=.85\linewidth]{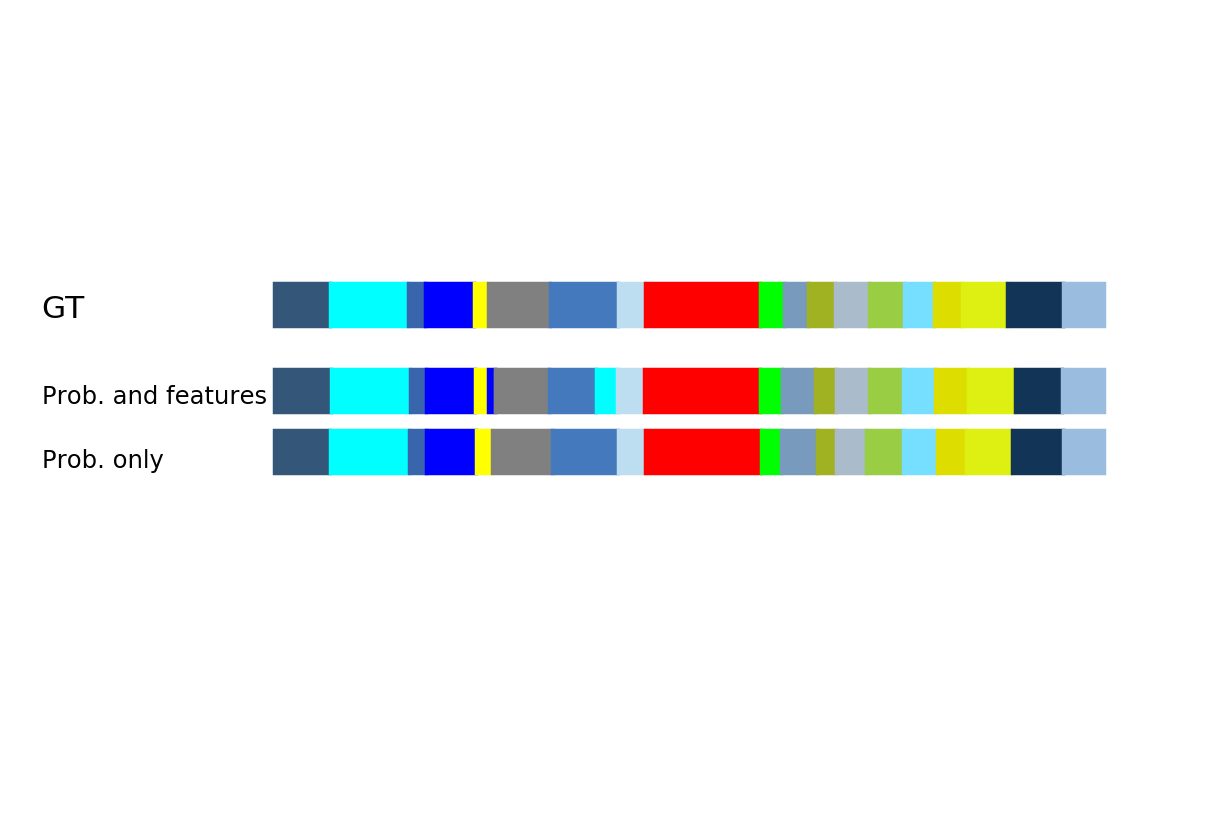}
\end{center}
   \caption{Qualitative results for two videos from the 50Salads dataset for showing the effect 
   of passing features to higher stages.}
\label{fig:features}
\end{figure}
\begin{figure*}[tb]
\begin{center}
\begin{tabular}{c}
   \includegraphics[trim={1cm 12cm 3cm 3.5cm},clip,width=.65\linewidth]{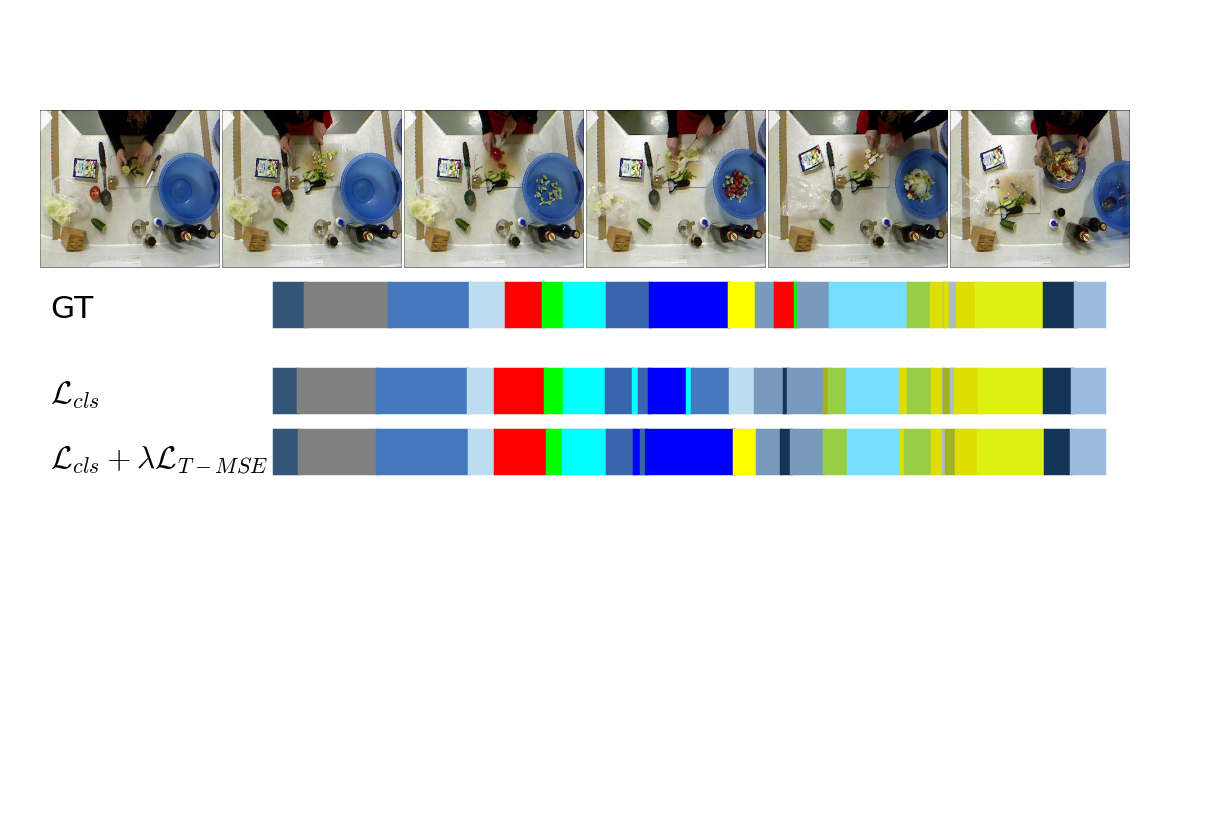}
  	\\
  	(a) 
  	\\
  	\includegraphics[trim={1cm 12cm 3cm 3.5cm},clip,width=.65\linewidth]{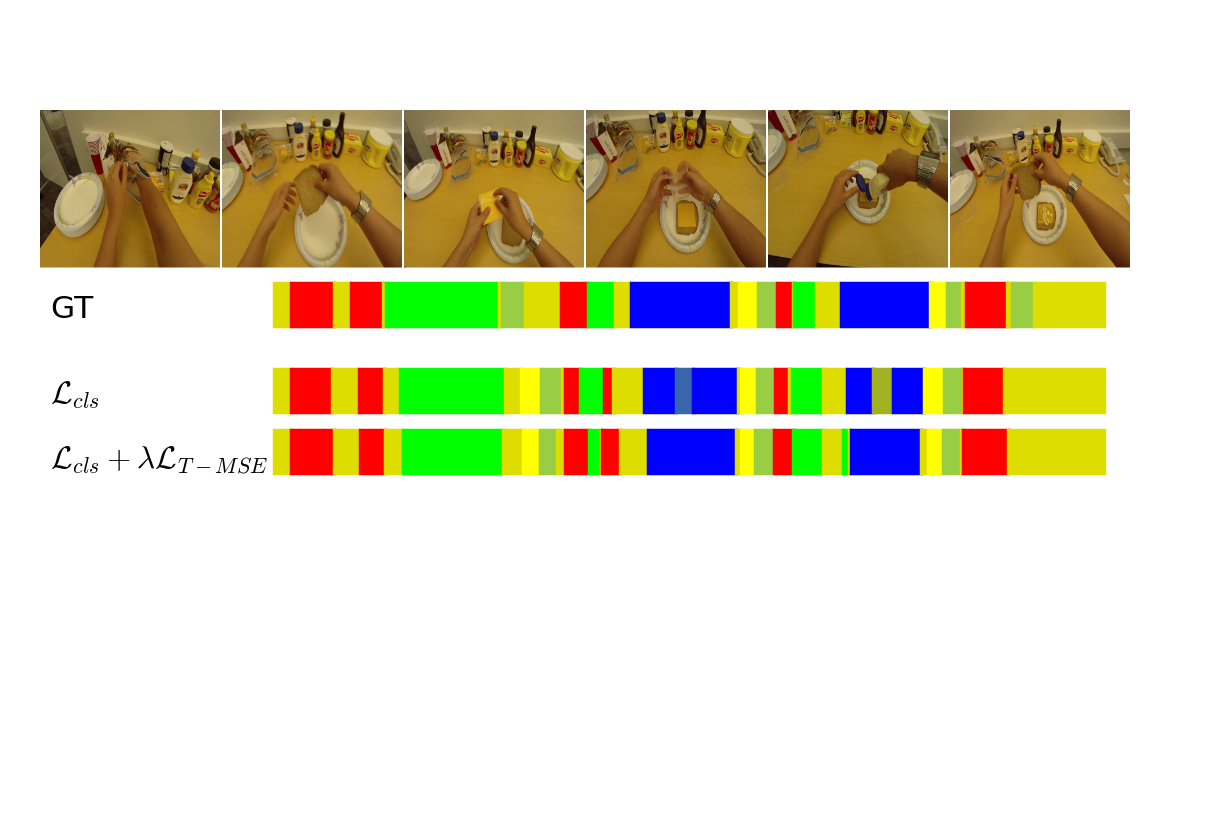}
  	\\
  	(b) 
  	\\
  	\includegraphics[trim={1cm 12cm 3cm 3.5cm},clip,width=.65\linewidth]{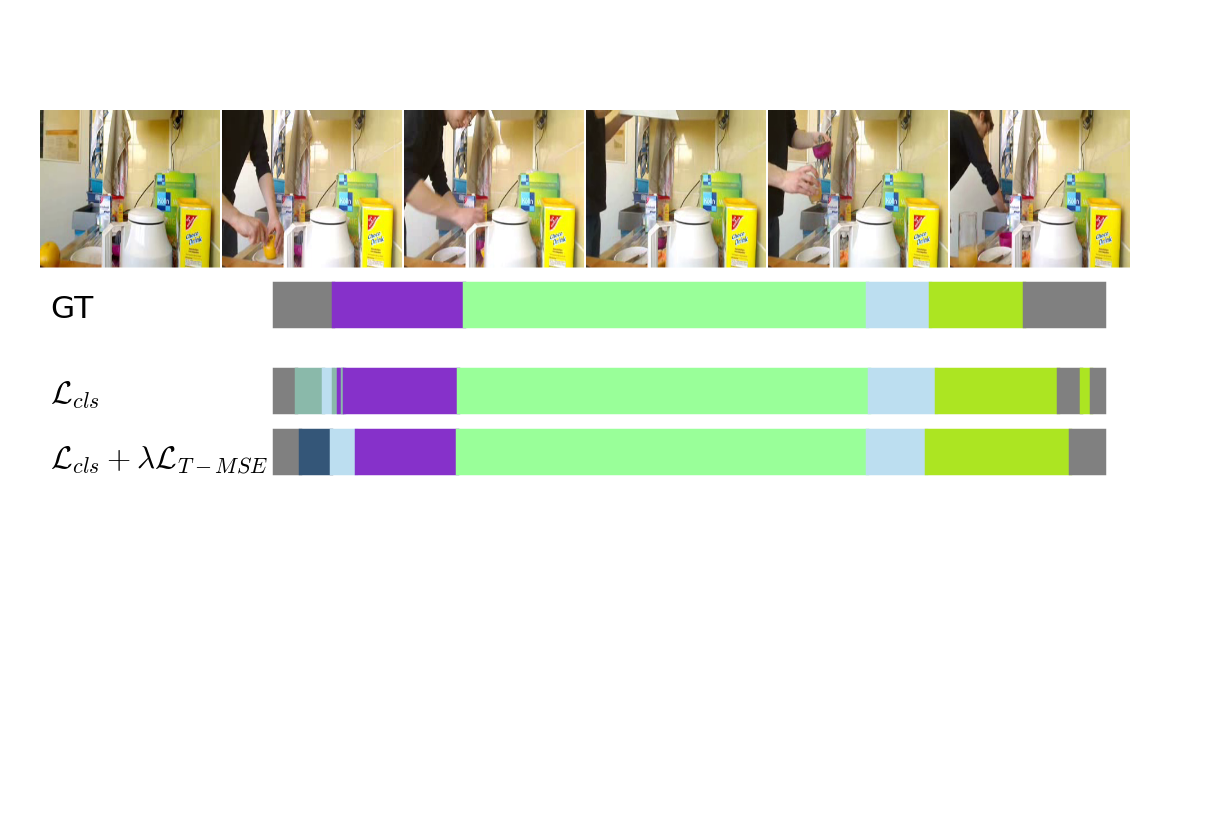}
  	\\
  	(c) 
\end{tabular}
\end{center}
   \caption{Qualitative results for the temporal action segmentation task on (a) 50Salads 
   (b) GTEA, and (c) Breakfast dataset.}
\label{fig:qualitative_res}
\end{figure*}
In the proposed multi-stage TCN, the input to higher stages are the 
frame-wise probabilities only. However, in the multi-stage architectures 
that are used for human pose estimation, additional features are 
usually concatenated to the output heat-maps of the previous stage.  
In this experiment, we therefore analyze the effect of combining additional 
features to the input probabilities of higher stages. To this end, we trained 
two multi-stage TCNs: one with only the predicted frame-wise probabilities 
as input to the next stage, and for the second model, we concatenated the output 
of the last dilated convolutional layer in each stage to the input probabilities 
of the next stage. As shown in Table~\ref{tab:features}, concatenating 
the features to the input probabilities results in a huge drop of the F1 score 
and the segmental edit distance (around $20\%$). We argue that the reason behind this 
degradation in performance is that a lot of action classes share similar appearance 
and motion. By adding the features of such classes at each stage, the model 
is confused and produces small separated falsely detected action segments 
that correspond to an over-segmentation effect. Passing only the probabilities 
forces the model to focus on the context of neighboring labels, which are explicitly 
represented by the probabilities. This effect can also be seen in the qualitative 
results shown in Figure~\ref{fig:features}.

\subsection{Impact of Temporal Resolution}

Previous temporal models operate on a low temporal resolution of 1-3 frames 
per second~\cite{Lea_2017_CVPR, lei2018temporal, ding2018weakly}. On the 
contrary, our approach is able to handle higher resolution of 15 fps. 
In this experiment, we evaluate our model in a low temporal resolution of 
1 fps. As shown in Table~\ref{tab:temporal_res}, the proposed model is able 
to handle both low and high temporal resolutions. While reducing the 
temporal resolution results in a better edit distance and segmental F1 score, 
using high resolution gives better frame-wise accuracy. Operating on a low 
temporal resolution makes the model less prune to the over-segmentation problem, 
which is reflected in the better edit and F1 scores. Nevertheless, this comes 
with the cost of losing the precise location of the boundaries between action 
segments, or even missing small action segments.

\begin{table}[tb]
\centering
\resizebox{.75\linewidth}{!}{%
\begin{tabular}{lccccc}
\hline
  & \multicolumn{3}{c}{F1@\{10,25,50\}} & Edit & Acc  
\\ \hline
MS-TCN (1 fps)     & \textbf{77.8} & \textbf{74.9} &         64.0  & \textbf{70.7} &         78.6  \\
MS-TCN (15 fps)    &         76.3  &         74.0  & \textbf{64.5} &         67.9  & \textbf{80.7} \\ 
\hline
\end{tabular}%
}
\caption{Impact of temporal resolution on the 50Salads dataset.}
\label{tab:temporal_res}
\end{table}

\subsection{Impact of the Number of Layers}
In our experiments, we fix the number of layers (L) in each stage to $10$ Layers. 
Table~\ref{tab:number_of_layers} shows the impact of this parameter on 
the 50Salads dataset. Increasing L form $6$ to $10$ significantly 
improves the performance. This is mainly due to the increase in the 
receptive field. Using more than $10$ layers (L = $12$) does not improve 
the frame-wise accuracy but slightly increases the F1 scores.

\begin{table}[tb]
\centering
\resizebox{.65\linewidth}{!}{%
\begin{tabular}{lccccc}
\hline
  & \multicolumn{3}{c}{F1@\{10,25,50\}} & Edit & Acc  
\\ \hline
L = 6      &         53.2  &         48.3  &         39.0  &         46.2  &         63.7  \\
L = 8      &         66.4  &         63.7  &         52.8  &         60.1  &         73.9  \\
L = 10     &         76.3  &         74.0  &         64.5  &         67.9  & \textbf{80.7}  \\ 
L = 12     & \textbf{77.8} & \textbf{75.2} & \textbf{66.9} & \textbf{69.6} &         80.5 \\ 
\hline

\end{tabular}%
}
\caption{Effect of the number of layers (L) in each stage on the 50Salads dataset.}
\label{tab:number_of_layers}
\end{table}

To study the impact of the large receptive field on short videos, we 
evaluate our model on three groups of videos based on their durations. 
For this evaluation, we use the GTEA dataset since it contains shorter videos 
compared to the others. As shown in Table~\ref{tab:impact_on_short_videos}, our 
model performs well on both short and long videos. Nevertheless, the performance is 
slightly worse on longer videos due to the limited receptive field.

\begin{table}[tb]
\centering
\resizebox{.7\linewidth}{!}{%
\begin{tabular}{lcccccc}
\hline
 Duration & \multicolumn{3}{c}{F1@\{10,25,50\}} & Edit & Acc  
\\ \hline
$< 1$ min        &         89.6  &         87.9  &         77.0  &         82.5  &         76.6  \\
$1 - 1.5$ min    &         85.9  &         84.3  &         71.9  &         80.7  &         76.4  \\
$\geq 1.5$ min   &         81.2  &         76.5  &         58.4  &         71.8  &         75.9  \\ 
\hline

\end{tabular}%
}
\caption{Evaluation of three groups of videos based on their durations on the GTEA dataset.}
\label{tab:impact_on_short_videos}
\end{table}

\subsection{Impact of Fine-tuning the Features}
In our experiments, we use the I3D features without fine-tuning. Table~\ref{tab:fine_tuning} 
shows the effect of fine-tuning on the GTEA dataset. Our multi-stage architecture 
significantly outperforms the single stage architecture - with and without fine-tuning. 
Fine-tuning improves the results, but the effect of fine-tuning for action segmentation is 
lower than for action recognition. This is expected since the temporal model is by far 
more important for segmentation than for recognition.

\begin{table}[ht]
\centering
\resizebox{.8\linewidth}{!}{%
\begin{tabular}{llccccc}
\hline
&  & \multicolumn{3}{c}{F1@\{10,25,50\}} & Edit & Acc  
\\ \hline
w/o FT   & SS-TCN            &         62.8  &         60.0  &         48.1  &         55.0  &   73.3        \\
         & MS-TCN (4 stages) & \textbf{85.8} & \textbf{83.4} & \textbf{69.8} & \textbf{79.0} & \textbf{76.3}        \\ 
\hline
with FT  & SS-TCN            &         69.5  &         64.9  &         55.8  &         61.1  &   75.3        \\
         & MS-TCN (4 stages) & \textbf{87.5} & \textbf{85.4} & \textbf{74.6} & \textbf{81.4} & \textbf{79.2}        \\
\hline

\end{tabular}%
}
\caption{Effect of fine-tuning on the GTEA dataset.}
\label{tab:fine_tuning}
\end{table}
%

\subsection{Comparison with the State-of-the-Art}
In this section, we compare the proposed model to the state-of-the-art 
methods on three datasets: 50Salads, Georgia Tech Egocentric Activities (GTEA), 
and Breakfast datasets. The results are presented in Table~\ref{tab:state_of_the_art}. 
As shown in the table, our model outperforms the state-of-the-art methods on the three 
datasets and with respect to three evaluation metrics: F1 score, segmental edit distance, 
and frame-wise accuracy (Acc) with a large margin (up to $12.6 \%$ for the frame-wise accuracy 
on the 50Salads dataset). Qualitative results on the three datasets are shown in 
Figure~\ref{fig:qualitative_res}. Note that all the reported results are obtained using the 
I3D features. To analyze the effect of using a different type of features, we evaluated our 
model on the Breakfast dataset using the improved dense trajectories (IDT) features, which are 
the standard used features for the Breakfast dataset. As shown in Table~\ref{tab:state_of_the_art}, 
the impact of the features is very small. While the frame-wise accuracy and edit distance are 
slightly better using the I3D features, the model achieves a better F1 score when using the IDT 
features compared to I3D. This is mainly because I3D features encode both motion and appearance, 
whereas the IDT features encode only motion. For datasets like Breakfast, using appearance 
information does not help the performance since the appearance does not give a strong evidence 
about the action that is carried out. This can be seen in the qualitative results shown in 
Figure~\ref{fig:qualitative_res}. The video frames share a very similar appearance. Additional 
appearance features therefore do not help in recognizing the activity.

As our model does not use any recurrent layers, it is very fast both during training and testing. 
Training our four-stages MS-TCN for $50$ epochs on the 50Salads dataset is four times faster than 
training a single cell of Bi-LSTM with a 64-dimensional hidden state on a single GTX 1080 Ti GPU. 
This is due to the sequential prediction of the LSTM, where the activations at any time step 
depend on the activations from the previous steps. For the MS-TCN, activations at all time steps 
are computed in parallel.

\begin{table}[tb]
\centering
\resizebox{.75\linewidth}{!}{%
\begin{tabular}{lccccc}
\hline
\textbf{50Salads} & \multicolumn{3}{c}{F1@\{10,25,50\}} & Edit & Acc  
\\ \hline
IDT+LM~\cite{richard2016temporal} &        44.4 &        38.9 &        27.8     &    45.8 &        48.7  \\ 
Bi-LSTM~\cite{singh2016multi}     &        62.6 &        58.3 &        47.0     &    55.6 &        55.7  \\
ED-TCN~\cite{Lea_2017_CVPR}       &        68.0 &        63.9 &        52.6     &    59.8 &        64.7  \\
TDRN~\cite{lei2018temporal}       &        72.9 &        68.5 &        57.2     &    66.0 &        68.1  \\ 
\hline
MS-TCN                       &\textbf{76.3}&\textbf{74.0}&\textbf{64.5}&\textbf{67.9}&\textbf{80.7} \\    
\hline
\hline
\textbf{GTEA} & \multicolumn{3}{c}{F1@\{10,25,50\}} & Edit & Acc  
\\ \hline
Bi-LSTM~\cite{singh2016multi} &         66.5  &         59.0  &         43.6  &          -    &         55.5  \\
ED-TCN~\cite{Lea_2017_CVPR}   &         72.2  &         69.3  &         56.0  &          -    &         64.0  \\
TDRN~\cite{lei2018temporal}   &         79.2  &         74.4  &         62.7  &         74.1  &         70.1  \\ 
\hline
MS-TCN                   &              85.8  &         83.4 &          69.8  &         79.0  &         76.3  \\  
MS-TCN  (FT)             &      \textbf{87.5} & \textbf{85.4} & \textbf{74.6} & \textbf{81.4} & \textbf{79.2}     \\ 
\hline
\hline
\textbf{Breakfast} & \multicolumn{3}{c}{F1@\{10,25,50\}} & Edit & Acc  
\\ \hline
ED-TCN~\cite{Lea_2017_CVPR}*  &          -    &          -    &          -    &          -    &         43.3  \\
HTK~\cite{kuehne2017weakly}   &          -    &          -    &          -    &          -    &         50.7  \\
TCFPN~\cite{ding2018weakly}   &          -    &          -    &          -    &          -    &         52.0  \\
HTK(64)~\cite{kuehne2016end}  &          -    &          -    &          -    &          -    &         56.3  \\   
GRU~\cite{richard2017weakly}* &          -    &          -    &          -    &          -    &         60.6  \\ 
\hline
MS-TCN (IDT)                  & \textbf{58.2} & \textbf{52.9} & \textbf{40.8} &         61.4  &         65.1  \\      
MS-TCN (I3D)                  &         52.6  &         48.1  &         37.9  & \textbf{61.7} & \textbf{66.3} \\      
\hline
\end{tabular}%
}
\caption{Comparison with the state-of-the-art on 50Salads, GTEA, and the Breakfast dataset. (* obtained from~\cite{ding2018weakly}).}
\label{tab:state_of_the_art}
\vspace{-4mm}
\end{table}

\section{Conclusion}

We presented a multi-stage architecture for the temporal action segmentation task. 
Instead of the commonly used temporal pooling, we used dilated convolutions to increase 
the temporal receptive field. The experimental evaluation demonstrated the capability of 
our architecture in capturing temporal dependencies between action classes and reducing 
over-segmentation errors. We further introduced a smoothing loss that gives an additional 
improvement of the predictions quality. Our model outperforms the state-of-the-art methods 
on three challenging datasets with a large margin. Since our model is fully convolutional, 
it is very efficient and fast both during training and testing.

\paragraph{Acknowledgements:} The work has been funded by the Deutsche Forschungsgemeinschaft (DFG, German Research Foundation) – GA 1927/4-1 (FOR 2535 Anticipating Human Behavior) and the ERC Starting Grant ARCA (677650).

{\small
\bibliographystyle{ieee_fullname}
\bibliography{egbib}
}

\end{document}